\documentclass[letterpaper, 10 pt, conference]{ieeeconf}  
\IEEEoverridecommandlockouts                              
\pdfoutput=1
\usepackage{graphics}
\usepackage{epsfig}     
\usepackage{textcomp}
\usepackage{times}      
\usepackage{amsmath}    
\usepackage{mathtools}  
\usepackage{amssymb}    
\usepackage{url}
\usepackage{commath}
\usepackage{graphicx}
\usepackage[table,xcdraw]{xcolor}

\usepackage{caption}
\usepackage{multirow}

\usepackage{subcaption}
\usepackage{array} 
\usepackage{url}
\usepackage{hyperref}
\hypersetup{colorlinks,allcolors=black}
\usepackage{matlab-prettifier}
\usepackage{siunitx}
\usepackage{xcolor}

\usepackage{listings}
\definecolor{codegreen}{rgb}{0,0.6,0}
\definecolor{codegray}{rgb}{0.5,0.5,0.5}
\definecolor{codepurple}{rgb}{0.58,0,0.82}
\definecolor{backcolour}{rgb}{0.95,0.95,0.92}

\lstdefinestyle{mystyle}{
    backgroundcolor=\color{backcolour},   
    commentstyle=\color{codegreen},
    keywordstyle=\color{magenta},
    numberstyle=\tiny\color{codegray},
    stringstyle=\color{codepurple},
    basicstyle=\ttfamily\footnotesize,
    breakatwhitespace=false,         
    breaklines=true,                 
    captionpos=t,                    
    keepspaces=true,                 
    numbers=left,                    
    numbersep=5pt,                  
    showspaces=false,                
    showstringspaces=false,
    showtabs=false,                  
    tabsize=2,
    frame=single
}

\DeclareCaptionFormat{listing}{\rule{\dimexpr\textwidth+17pt\relax}{0.4pt}\par\vskip1pt#1#2#3}

\usepackage{hyperref}
\hypersetup{
colorlinks=true,
linkcolor=blue,
filecolor=magenta,
urlcolor=blue,
}

\usepackage[font=footnotesize]{caption}

\title{\LARGE \bf
Implications of Personality on Cognitive Workload, Affect, and Task Performance in Remote Robot Control
}

\author{Go-Eum Cha, Wonse Jo, and Byung-Cheol Min
\thanks{This material is based upon work supported by the National Science Foundation under Grant No. IIS-1846221. Any opinions, findings, and conclusions or recommendations expressed in this material are those of the author(s) and do not necessarily reflect the views of the National Science Foundation.}
\thanks{All authors are with SMART Lab, Department of Computer and Information Technology, Purdue University, West Lafayette, IN 47907, USA \tt\small{[cha20, jow, and minb] @purdue.edu}}%
}

\begin{document}
\maketitle
\thispagestyle{empty}
\pagestyle{empty}

\begin{abstract} 

This paper explores how the personality traits of robot operators can influence their task performance during remote control of robots. It is essential to explore the impact of personal dispositions on information processing, both directly and indirectly, when working with robots on specific tasks. To investigate this relationship, we utilize the open-access multi-modal dataset MOCAS to examine the robot operator's personality traits, affect, cognitive load, and task performance. Our objective is to confirm if personality traits have a total effect, including both direct and indirect effects, that could significantly impact the performance levels of operators. Specifically, we examine the relationship between personality traits such as extroversion, conscientiousness, and agreeableness, and task performance. We conduct a correlation analysis between cognitive load, self-ratings of workload and affect, and quantified individual personality traits along with their experimental scores. The findings show that personality traits do not have a total effect on task performance. A supplementary video can be accessed at: \href{https://youtu.be/h3XUtVn7nzg}{https://youtu.be/h3XUtVn7nzg}.
\end{abstract}

\section{Introduction}
\label{sec:introduction}

Continued advancements in robotics have enabled remote control applications, such as accessing places that are inaccessible to robot operators \cite{chiou2022robot} and surveillance \cite{dadashi2013semi}. These tasks typically involve repetition and require continuous human perception. Human factors in remote control, such as cognitive workload, trust, stress, and affect \cite{hopko2022human, baltrusch2022human}, have been commonly identified as decisive factors in achieving better task performance. These factors can be classified as intrinsic and extrinsic factors in a given collaboration environment and can be measured to evaluate the operator's task performance. Intrinsic factors, such as human trust in robots \cite{hancock2011meta} and recognition of formed relationships \cite{charalambous2015identifying}, as well as external factors such as the speed of the robot's movement \cite{koppenborg2017effects}, have considerable influences on the collaborative relationship with robots, as determined by collective variables. 

\begin{figure}[t]
    \centering
    \includegraphics[width=0.95\linewidth]{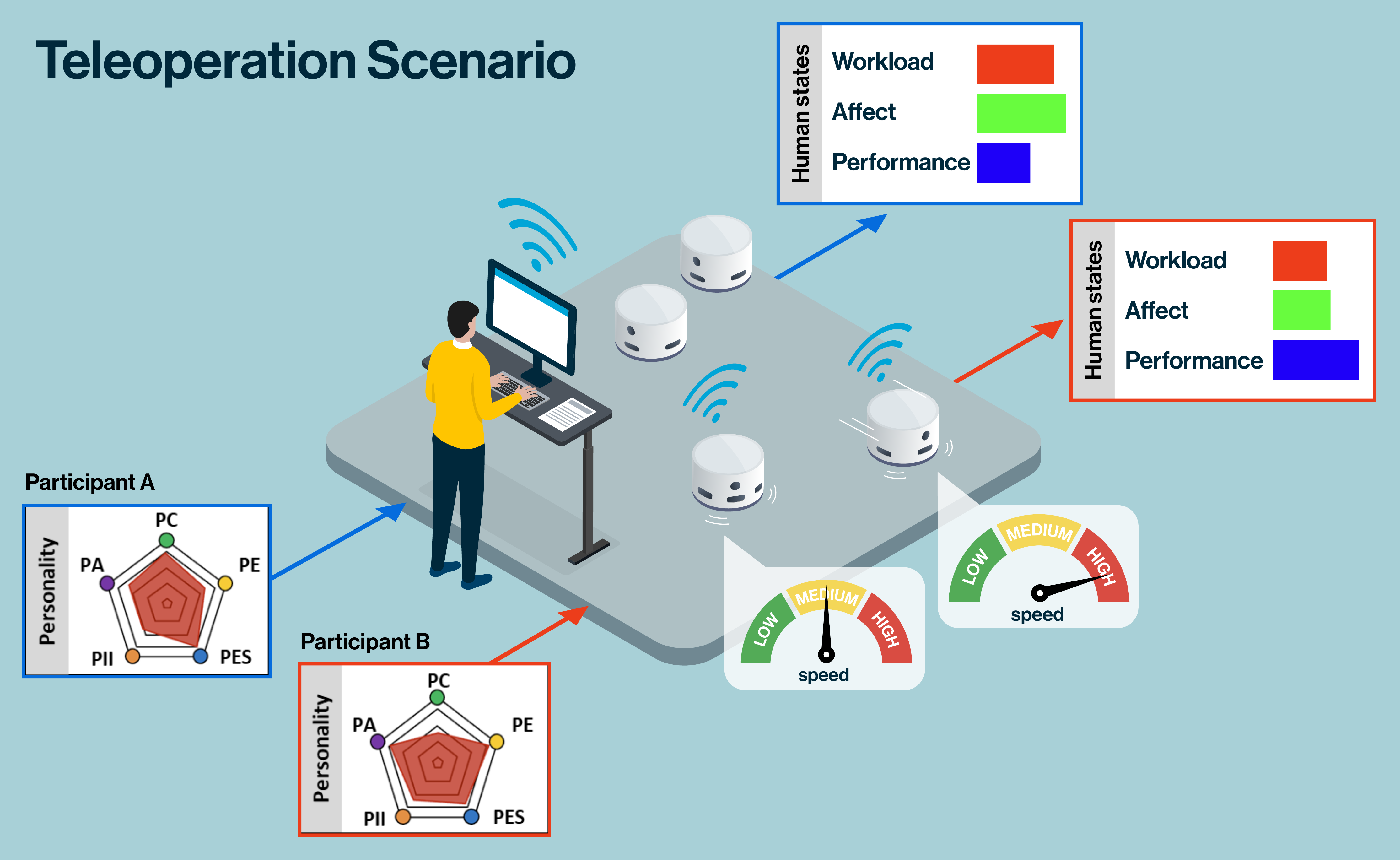} 
    \caption{Illustration of the influence of personality on human states (workload, affect, and performance) during remote robot control. Each label indicates as following: Extroversion (PE), Emotion stability (PES), Agreeableness (PA) Conscientiousness (PC), and Imagination/Intellect (PII).}
    \label{fig:concept}
    \vspace{-15pt}
\end{figure}

Task performance in video-based surveillance or monitoring systems, where human operators are obligated to remotely control robots, can be significantly impacted by factors related to both robots and humans \cite{dadashi2013semi, saini2012adaptive}, as illustrated in Fig.~\ref{fig:concept}. Operators are mainly responsible for taking action, such as reporting abnormalities (e.g., violent incidents) when they are discovered while monitoring crowded public spaces. Extrinsic variables, such as the complexity of tasks and accuracy of the system, and the information obtained from the multiple cameras can exert intrinsic effects on the operators, such as mental workload and affective states, which are directly related to their task performance \cite{dadashi2013semi}. 

The role of implicit perception in information processing and behavior is strongly linked to personality traits and can vary depending on the individual \cite{humphreys1984personality, hoyle2006personality}. 
Meta-analyses in multiple disciplines, such as psychology and human-robot interaction, have demonstrated that personality traits play significant roles in processing information and forming relationships that are not solely dictated by independent factors. Multiple contextual elements, such as personal disposition, situational and task-related context, and affect, may jointly influence human perception \cite{humphreys1984personality}. Additionally, the association of multiple human-related characteristics, such as personality traits or attitudes towards robots, also shows their relevance in developing rapport between human and robot entities \cite{hancock2011meta}.

Human factors have been analyzed in the context of human temporal conditions, but personal inclinations may also influence task performance \cite{baltrusch2022human}. Personality traits, as one of the intrinsic human factors, refer to a set of personal characteristics that have been empirically shaped over time. Psychologists or social scientists have taxonomically labeled these traits  \cite{goldberg1990alternative, ashton2014hexaco}. These traits have been mainly analyzed to determine which types show better performance in given tasks \cite{piepiora2021assessment, poropat2009meta}. For example, the personalities of athletes are distributed differently according to the type of team sports and also affect the team's performance \cite{piepiora2021assessment}. However, the available literature has not extensively discussed the dependence between dispositions and measurable task elements, although the implications of personality on social acceptance \cite{esterwood2021meta, santamaria2017personality} and trust \cite{salem2015would} have been discussed.  

The main objective of this paper is to investigate the impact of operator's personality on their cognitive workload, affect, and task performance in remote control situations. Additionally, we aim to determine the extent to which personality traits affect task performance with varying temporal states. To accomplish this, we use our open-access multi-modal MOCAS dataset \cite{jo2022mocas}, which contains data collected from human subjects conducting a surveillance task using Closed-Circuit Television (CCTV) systems. We conduct extensive statistical analysis to investigate the relevance of personality traits to task performance and the behavioral characteristics of operators through their cognitive and affective responses.

The main contributions of our work are the following:
\begin{itemize}
    \item We statistically investigate correlations of personal dispositions, cognitive workload, affect, and task performance from human participants;
    \item We verify the existence of direct and indirect effects between those measurements; 
    \item We present a comprehensive discussion of personality in human-robot teams with quantitative and qualitative studies. 
\end{itemize}

\section{Background}
\label{sec:background}

Personality traits have been studied in two separate contexts: traits of human users and robots. Personal disposition in individuals can be measured  through personality models such as the Big-Five Factor Models (FFM) \cite{goldberg1990alternative} and HEXACO \cite{ashton2014hexaco}. Additionally, robots can be programmed to exhibit certain personality traits \cite{santamaria2017personality}, which may affect how similar humans perceive them. For example, researchers have found that personality traits influence human participants' preference \cite{tapus2008user} and acceptance \cite{esterwood2021birds}, matching their personality projected from robots, from pilot studies to meta-analyses. While the association between robot personality and human perception is important \cite{santamaria2017personality}, the authors specifically focus on human disposition in remote control scenarios, particularly in relation to surveillance tasks, where robots display minimal predetermined personality traits.

Personality traits, more specifically the introversion-extroversion dimension, have a prominent association with task completion from a psychological perspective. While this personality dimension may not be directly dependent on perception, it has a robust relationship with impulsivity \cite{humphreys1984personality, hoyle2006personality}. Research has suggested that impulsivity can serve as an intrinsic stimulus, motivating individuals to achieve better performance outcomes. Typically, impulsivity is oppositely associated with the dimension of conscientiousness presenting that self-regulation is attributed to higher conscientiousness \cite{hoyle2006personality}. 
The magnitude of diligence empirically showed their carefulness in differentiating abnormalities (errors) in a short duration which resulted from the degree of self-discipline \cite{hill2016contextualizing}. The higher diligence also seems to be translated into higher work engagement resulting in better task performance \cite{bakker2012work}.
Contrary to carefulness that higher diligent individuals may have, persons higher on agreeableness tend to commit errors in given tasks supported by empirical evidence that people who have higher conscientiousness outperformed those who have higher agreeableness in tasks of self-regulation \cite{jensen2002agreeableness}. The result is potentially supported that more optimistic people tend to demonstrate rapid decision seen in a visual-matching problem \cite{jensen2002agreeableness}. The previously mentioned studies prove that individuals with higher agreeableness could show higher error rates in tasks.

Experts also contend that the level of extroversion is associated with individuals' socio-psychological and physiological reactions, as well as their task performance during completion. For example, the Yerkes--Dodson law \cite{yerkes1907dancing}, which is represented by a bell-shaped curve, shows an empirical relationship between stimuli and performance. The law suggests that performance initially increases with the level of arousal, but then drops off rapidly at a certain point, known as the apex, at which the level of arousal persists. 

Empirical research has shown a link between stimuli, information processing, and task performance. However, due to the inconsistent evaluation of personality traits, available studies also demonstrate discrepancies between them. Some studies found that extroversion had no discernible impact on task performance \cite{dang2015stress, lee2023effect}, which is inconsistent with previously discussed studies. 
Similarly, a study in \cite{dang2015stress} also found that the level of extroversion exhibited by participants impacted the frequency of errors that could arise during collaboration with a robot. However, this finding is inherently challenging to apply directly to task performance, given that the experiment focused more on the relationship between robots and operators \cite{dang2015stress}. Considering the effects of personality traits in \cite{humphreys1984personality}, the effect on task completion could be entirely originated from not only personality traits but also the level of arousal and effort.
Therefore, additional investigation may be necessary to fully understand the complexities of the effect of personality traits in remote robot control scenarios. 


\subsection{Hypotheses}

This study aims to investigate the impact of personality traits on task performance by examining the relationship between cognitive and emotional fluctuations resulting from information processing. We hypothesize that:

\begin{itemize}
    \item $\mathcal{H}_1$: Each item of FFM has a positive or negative correlation with task completion. Participants with higher levels of extroversion show higher task performance ($\mathcal{H}_1$a), as do participants with higher levels of conscientiousness ($\mathcal{H}_1$b). 
    \item $\mathcal{H}_2$: The success performance rate of operators may be correlated with individual personality traits. Participants with higher levels of conscientiousness have a higher success rate and task performance ($\mathcal{H}_2$a), while participants with higher levels of agreeableness have a lower success rate and task performance ($\mathcal{H}_2$b). 
    \item $\mathcal{H}_3$: Subjective assessments of cognitive workload and affect may be related to personality traits. Participants with higher levels of extroversion show higher cognitive workload and higher arousal simultaneously ($\mathcal{H}_3$a).
    \item $\mathcal{H}_4$: Personality traits may have both direct and indirect effects on task performance. 
    
\end{itemize}


\section{Dataset} 
\label{sec:design_experiment}
\begin{figure}[t]
    \centering
    \begin{subfigure}{0.61\linewidth}
        \centering
        \includegraphics[width=1\linewidth]{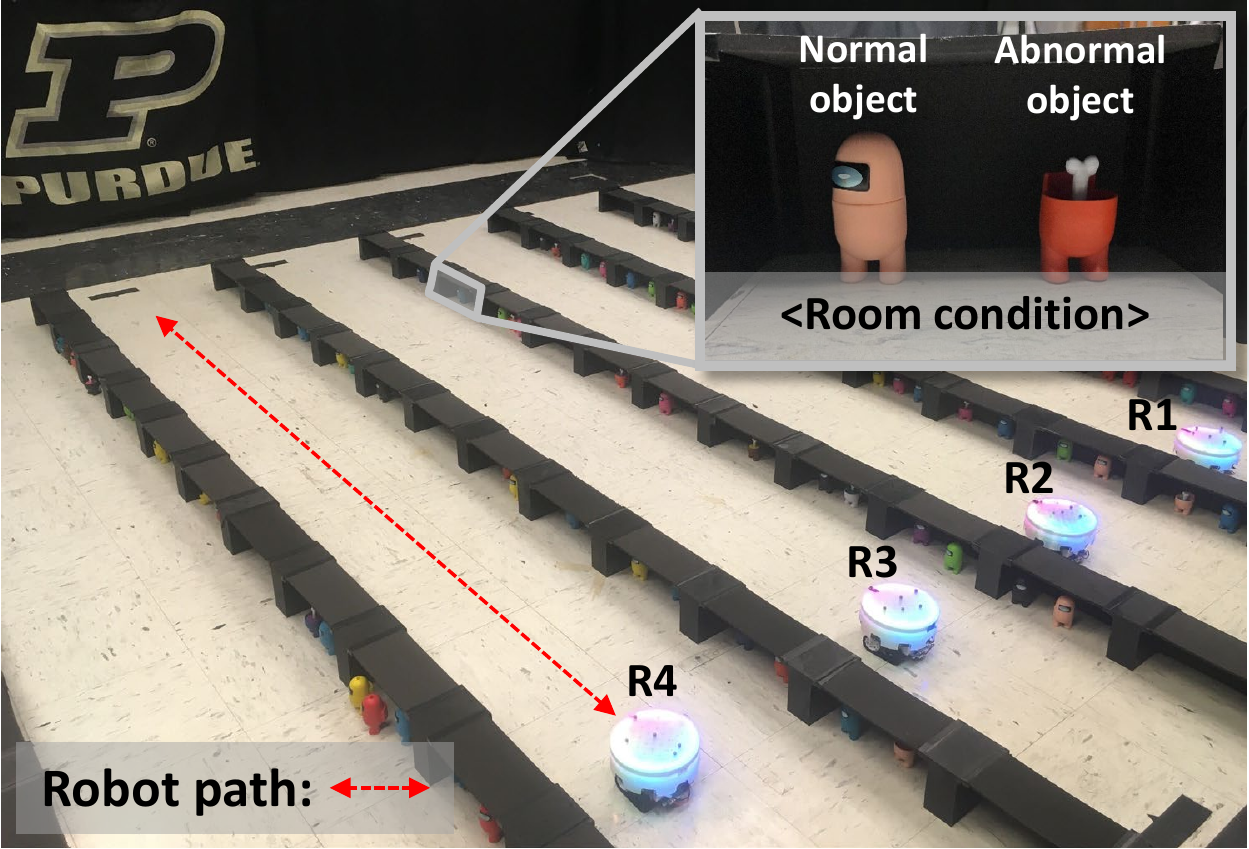}
        \caption{}
        \label{fig:mrs_testbed}
    \end{subfigure}
    \begin{subfigure}{0.34\linewidth}
        \centering
        \includegraphics[width=1\linewidth]{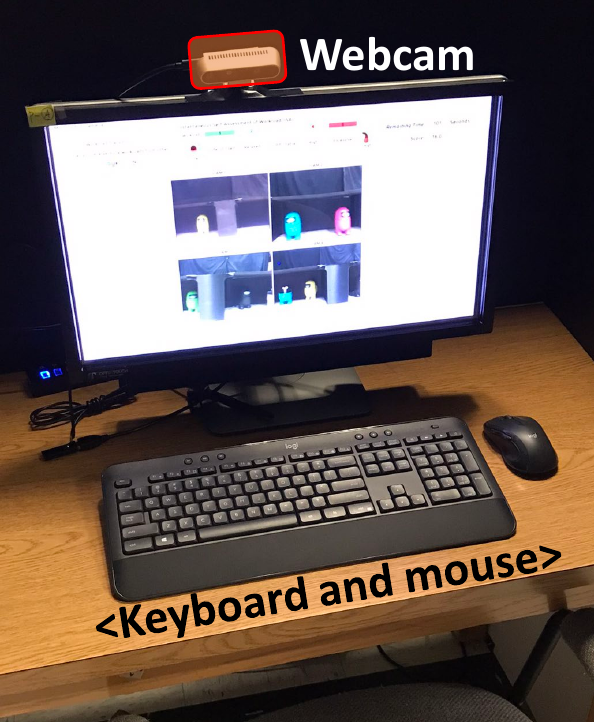}
        \caption{}
        \label{fig:human_testbed} 
    \end{subfigure}
    
    \caption{Experiment environment used for building the MOCAS dataset; 
    (a) robot testbed and a room condition in which normal and abnormal objects are located (shown in the gray box), and (b) details of the human subject's desk used to remotely conduct the CCTV monitoring mission.
    }
    \label{fig:mocas_environemt}
    \vspace{-15pt}
\end{figure}

For this study, we utilize our open-access multi-modal dataset, called \textit{MOCAS} \cite{jo2022mocas}\footnote{The MOCAS dataset is open-access, and further details about it can be found at \url{https://polytechnic.purdue.edu/ahmrs/mocas-dataset}.}, that contains data from 21 participants, including physiological signals, facial camera videos, mouse movements, screen recordings, and subjective questionnaires based on different levels of cognitive load \cite{jo2022mocas}. 
The participant's ages ranged from 18 to 37 years old (mean = 24.3 years, S.D. = 5.2 years).
The stimulus used in the dataset is a CCTV monitoring task, in which a human operator monitors single or multiple camera views streamed by multiple patrol robot platforms. 
The number and speed of robots could be adjusted to present varying levels of stimuli, such as low, medium, and high workloads. 
During the CCTV monitoring task, participants were tasked with detecting and clicking abnormal objects (see the red-dashed box in Fig.~\ref{fig:mocas_environemt}) from the camera views. Clicking on an abnormal object correctly earned 1 point while failing to do so resulted in a loss of 3 points.

\subsection{Procedure}

\begin{figure}[t]
    \centering
    \includegraphics[width=0.85\linewidth]{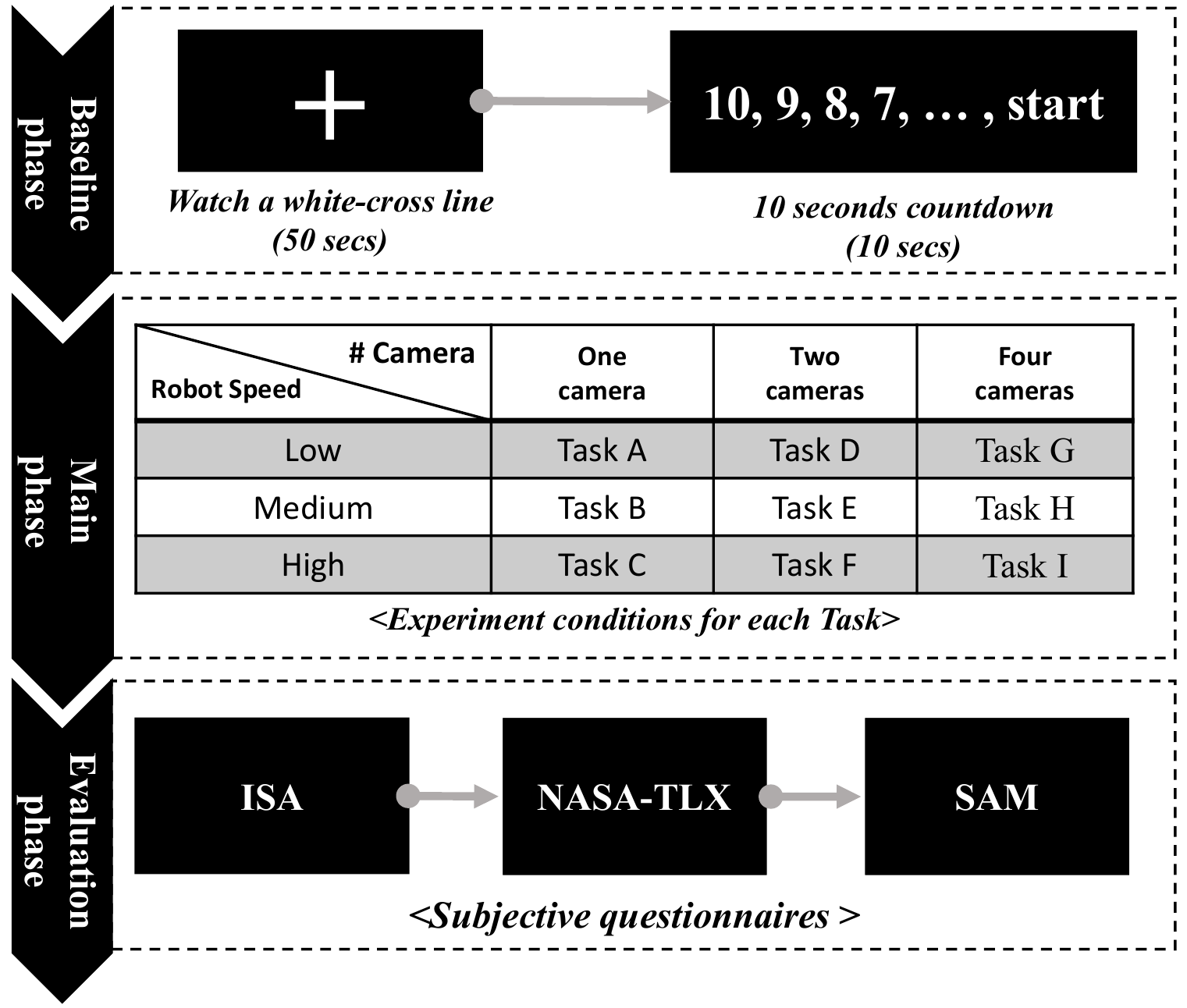}
    \caption{Overall procedures for the data collection used in the MOCAS dataset.}
    \label{fig:overall_procedures}
    \vspace{-20pt}
\end{figure}

Fig.~\ref{fig:overall_procedures} illustrates the overall data collection procedures used in the MOCAS dataset. 
Before starting the experiment, the participants were required to go through the informed consent, and complete demographic and personality questionnaires which the IPIP Big-Five Factor Markers to categorize personality traits into five distinct categories \cite{goldberg1990alternative}.

The main experiment has three repeated phases as illustrated in the dashed box in Fig.~\ref{fig:overall_procedures}. 
The baseline phase is the preparation step to start the main phase. 
The main phase is for the participant to conduct the single CCTV monitoring task with different workload levels made by different combinations of three distinct numbers of camera views (e.g., one, two, or four cameras) and three different speeds of the multi-robot system (e.g., low, medium, and high speed) as presented in a table in Fig.~\ref{fig:overall_procedures}. 
The evaluation phase is to collect subjective cognitive load via Instantaneous Self-Assessment (ISA) \cite{jordan1992instantaneous} and NASA-Task Load Index (NASA-TLX) \cite{hart1988development} measures, as well as their subjective emotional state using the Self-Assessment Manikin (SAM) \cite{bradley1994measuring}.
Additionally, there is a supplementary video at \url{https://youtu.be/BxVVj7R9b70} that explains the details of this experimental design and procedures used for building the MOCAS dataset.

\subsection{Data collection} 
\subsubsection{Personality traits}
The MOCAS dataset includes the participants' FFM information collected from them before starting the main experiment. This paper considers five personality factors, which are described as follows.

\begin{itemize} 
    \item Extroversion (PE): A participant who scores high on the test typically exhibits outgoing, talkative, and social behavior, whereas those with low scores tend to demonstrate a more reserved and introspective demeanor.
    \item Emotional stability (or neuroticism; PES): A participant who scores high on the test tends to have a sensitive and nervous disposition, whereas someone with low scores is generally more resilient and confident in their behavior. 
    \item Agreeableness (PA): A participant who scores high on the test is inclined towards being friendly and optimistic, while those who score low are more likely to be critical and aggressive.
    \item Conscientiousness (PC): A participant who scores high on the test tends to display careful and diligent behavior, while those with low scores tend to exhibit impulsive and disorganized tendencies.
    \item  Intellect/Imagination (or openness to experience; PII): A participant who scores high on the test tends to be inventive and curious, whereas one who has low scores tends to be traditional and conventional. 
\end{itemize}

\subsubsection{Performance}
There are two variables to measure the participant's mission performance; \textit{Mission\_Score} and \textit{Success\_Rate}.
\begin{itemize}
    \item \textit{Mission\_Score}: the sum of the points obtained during the CCTV monitoring task and is referred to as the human operator's mission performance.
    \item \textit{Success\_Rate}: referred to as the quality of the participant's mission which is obtained by dividing the number of success clicks and total mouse clicks.
\end{itemize}

\subsubsection{Experiment}
There are two experimental factors used in the CCTV monitoring mission of the MOCAS dataset; 
\textit{Camera\_Number} and \textit{Robot\_Speed}. 
\begin{itemize}
    \item \textit{Camera\_Number}: assigned workload for human operators to monitor the number of camera views simultaneously. The range of {Camera\_Number} has three different numbers of camera views; one, two, or four cameras. 
    \item \textit{Robot\_Speed}: the speed of the robot platform for conducting patrol missions. The \textit{Robot\_Speed} has three different speeds of the robot platform; low (100 $mm/s$), medium (200 $mm/s$), and high speed (300 $mm/s$). 
\end{itemize}

\subsubsection{Questionnaire}
There are two variables to measure participants' cognitive workload through two self-reporting assessment tools; \textit{ISA} and \textit{Weighted\_NASA}.
\begin{itemize}
    \item \textit{ISA}: self-reported stress levels directly reported by the participant after finishing each task having different combinations of experimental factors (e.g., robot speed and camera number). 
    \item \textit{Weighted\_NASA}: a modified raw NASA-TLX by multiplying weights ($ [5, 0, 4, 3, 2, 1]$) on each NASA-TLX factor to provide a more accurate and personalized assessment of workload \cite{sugarindra2017mental}. 
\end{itemize}

\section{Statistical Investigation}
\label{sec:analysis}
The MOCAS dataset was divided into nine levels based on the three types of robot speed (e.g., 100 $mm/s$, 200 $mm/s$, and 300 $mm/s$) and the number of cameras (e.g., 1, 2, and 4). 
The segmented data of the 21 participants were 187 instead of 189 because one of the participants (P2) could not complete two segmented tasks (the combination of one camera$\times$200 $mm/s$ and two cameras$\times$100 $mm/s$). Table~\ref{tab:descstat} represents descriptive statistics to investigate the effect of measured information in the experiment. Statistical analysis was performed by IBM SPSS 29.0 for Windows. 

\begin{table}[h]
    \centering
    \caption{A table of descriptive statistics.}
    \label{tab:descstat}
    \resizebox{0.90\linewidth}{!}{%
    \begin{tabular}{lcccc}
        \hline
         & Min. & Max. & Mean & Std. \\ \hline
        \textbf{Performance} & \multicolumn{4}{l}{} \\ \hline
        Scores & -65 & 164 & 70.3 & 40.196 \\
        Success Rates & 0 & 1 & 0.94 & 0.076 \\ \hline
        \textbf{Personality} & \multicolumn{4}{l}{} \\ \hline
        Extroversion & 16 & 96 & 58.26 & 26.974 \\
        Emotional Stability & 4 & 93 & 50.11 & 23.896 \\
        Agreeableness & 1 & 89 & 49.53 & 26.133 \\
        Conscientiousness & 0 & 93 & 59.02 & 24.661 \\
        Intellect/Imagination & 6 & 93 & 53.8 & 28.364 \\ \hline
        \textbf{Subjective Ratings} & \multicolumn{4}{l}{} \\ \hline
        Weighted NASA-TLX & 13 & 90 & 47.56 & 18.242 \\
        ISA & -2 & 2 & 0.02 & 0.975 \\
        Arousal & -4 & 4 & 1.28 & 2.023 \\
        Valence & -4 & 4 & -0.65 & 2.494 \\ 
        \hline
        \end{tabular}
    }%
    \vspace{-5pt}
\end{table}

The rmANOVA results presented in \cite{jo2022mocas} indicate statistically significant differences in participant's cognitive loads based on robot speed ($F(2, 38) = 30.84, p < .001, \eta_{p}^{2} = 0.62$) and the number of camera views ($F(2, 38) = 36.3, p < .001, \eta_{p}^{2} = 0.66$), and the normality of data is satisfied. The results indicate that experimental variables successfully derived different levels of cognitive loads.

Pearson's correlation coefficient ($\gamma$) shows positive or negative relationships between experimental and performance variables, as represented in Table~\ref{tab:pearson_score1}. We followed the normative guidelines for interpreting correlation coefficients presented by the authors in \cite{gignac2016effect} to determine the magnitude of effect size.

\begin{table}[h]
\caption{Pearson’s correlation coefficients between score, success rate, the number (\#) of successful clicks, \# of cameras, and the speed of robots.}
\label{tab:pearson_score1}
\resizebox{\columnwidth}{!}{%
\begin{tabular}{lccccc}
\hline
\textbf{} &
  \multicolumn{1}{l}{\textbf{Score}} &
  \multicolumn{1}{l}{\textbf{Success Rate}} &
  \multicolumn{1}{l}{\textbf{Success Clicks}} &
  \multicolumn{1}{l}{\textbf{Camera \#}} &
  \multicolumn{1}{l}{\textbf{Speed}} \\ \hline
\textbf{Score}          & 1        &          &          &       &   \\
\textbf{Success Rate}   & .683***  & 1        &          &       &   \\
\textbf{Success Clicks} & .759***  & .072     & 1        &       &   \\
\textbf{Camera \#}      & .122     & -.364*** & .497***  & 1     &   \\
\textbf{Speed}          & -.582*** & -.611*** & -.304*** & -.002 & 1 \\ \hline
\footnotesize{ $^{***}$: $p$ \textless $.001$ }
\end{tabular}
\vspace{-60pt}
}
\end{table}

Overall, the success rate of participants showed a positive relationship with score ($\gamma$=.683, $p$ \textless $.001$) and negative relationships with the number of cameras ($\gamma$=-.364, $p$ \textless $.001$) and speed ($\gamma$=-.611, $p$ \textless $.001$). Also, the speed of robots showed a significant negative relationship with score ($\gamma$=-.582, $p$ \textless $.001$). The result indicates that the success click rates show gradual relevancy in measurement. The number of successful clicks of participants was strongly influenced by the number ($\gamma$=.497, $p$ \textless $.001$) and speed of cameras ($\gamma$=-.304, $p$ \textless $.001$), which means that the higher the level of difficulty, the lower the success rate of clicks. Moreover, speed has a negative impact on scores; allegedly, the speed of robots is the robust factor influencing performance. In addition, how successfully the participants clicked during the experiment may be a factor in determining the score, which is caused by penalties when incorrect clicks were attempted. 

\begin{figure}[t]
    \centering
    \includegraphics[width=0.90\linewidth]{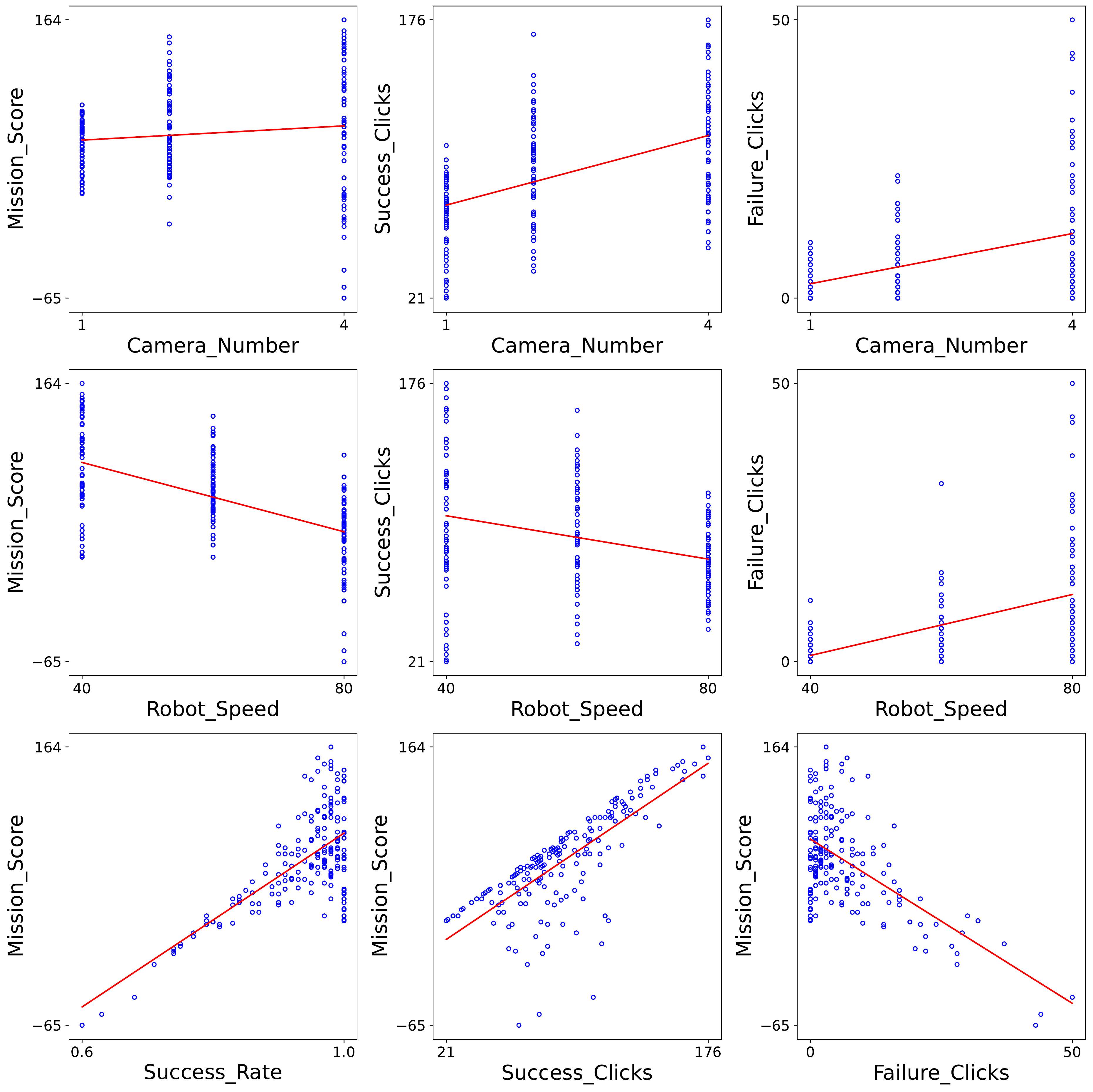}
    \caption{Scatterplots and least-square linear fit pairs between the number of cameras (Camera\_Number), the speed of robots (Robot\_Speed), mission scores, the rate of successful clicks, the number of successful clicks, and failure ones.}
    \label{fig:overall_experimental_data}
\end{figure}

Data in Fig.~\ref{fig:overall_experimental_data} illustrates scatter plots of partial pairs between experimental variables. Multiple outliers are identified in each plot, and no discernible non-linear relationships are observed. Consequently, we conclude that experimental variables alone are not sufficient to determine temporal human features, such as workload and performance.

\subsection{Variable Assessment with Personality}

\subsubsection{Pearson's correlation} 
Table~\ref{tab:pearson_score} presents the relationship between personal trait indicators and experimental and performance variables. Our findings partially support the expected relationship between personality indicators and higher performance ($\mathcal{H}_1$) when the dimension value of extroversion is higher ($\mathcal{H}_1$a, $\gamma$ = .148, $p$ \textless $.05$), while we did not find a significant relationship with conscientiousness ($\mathcal{H}_1$b). We also found that agreeableness has a small negative influence on the success click rate ($\gamma$ = -.153, $p$ \textless $.05$), while its score measurement did not show significance ($\gamma$ = -.124, $p$ \textgreater $.05$; $\mathcal{H}_2$b). Other indicator did not show their significance in deciding the performance measures. Therefore, we did not find a significant relationship between success click rate and conscientiousness ($\mathcal{H}_2$a). Following, SAM ratings separately collected valence and arousal showed significance in the pairs of PII-Valence ($\gamma$ = .243, $p$ \textless $.01$), PES-Valence ($\gamma$ = -.168, $p$\textless $.05$), and PA-Arousal ($\gamma$ = -.162, $p$ \textless $.01$). The strongest correlation showed in the pair of PE-Arousal, where the coefficient value is the highest ($\gamma$ = .335, $p$ \textless $.01$). The correlation between subjective assessments regarding cognitive workload and affect ($\mathcal{H}_3$) had more evidence than other previous hypotheses. For example, people with higher extroversion showed higher arousal and higher workload  simultaneously ($\mathcal{H}_3$a). The objective and subjective workload indicators showed their significance (Weighted NASA-TLX: $\gamma$ = .234, $p$ \textless $.01$, ISA: $\gamma$ = .196, $p$ \textless $.01$).

\begin{table}[t]
    \caption{Pearson's correlation coefficient between personality traits, performance, experimental indicators, SAM, Weighted NASA--TLX, and ISA.}
    \label{tab:pearson_score}
    \begin{center}
    \resizebox{0.95\columnwidth}{!}{%
    \begin{tabular}{llccc}
    \hline
 & \textbf{Score} & \textbf{Success Rate} & \textbf{Valence}  \\ \hline
    \textbf{PE} & .148$^{*}$ & -.024 & -.054 \\
    \textbf{PES} & -.023 & -.018 & -.168$^{*}$ \\
    \textbf{PA} & -.124 & -.153$^{*}$ & -.082 \\
    \textbf{PC} & .058 & -.044 & .086 \\
    \textbf{PII} & -.128 & -.015 & .243$^{**}$ \\
    \hline
 & \textbf{Arousal} & \textbf{Weighted NASA-TLX} & \textbf{ISA} \\ \hline
 \textbf{PE} & .335$^{**}$ & .234$^{**}$ & .196$^{**}$ \\
    \textbf{PES} & .059 & .159$^{*}$ &.215$^{**}$ \\
    \textbf{PA} & -.162$^{*}$ & -.170$^{*}$ & -.051 \\
    \textbf{PC} & .083 & .162$^{*}$ & .247$^{**}$ \\
    \textbf{PII} & -.046 & .014 & -.066\\
    \hline
    \footnotesize{$^{**}$: $p$ \textless $.01$,  $^{*}$: $p$ \textless $.05$}
    \end{tabular}
    }%
    \end{center}
    \vspace{-20pt}
\end{table}

\begin{figure*}[h]
    \centering
    \includegraphics[width=0.98\linewidth]{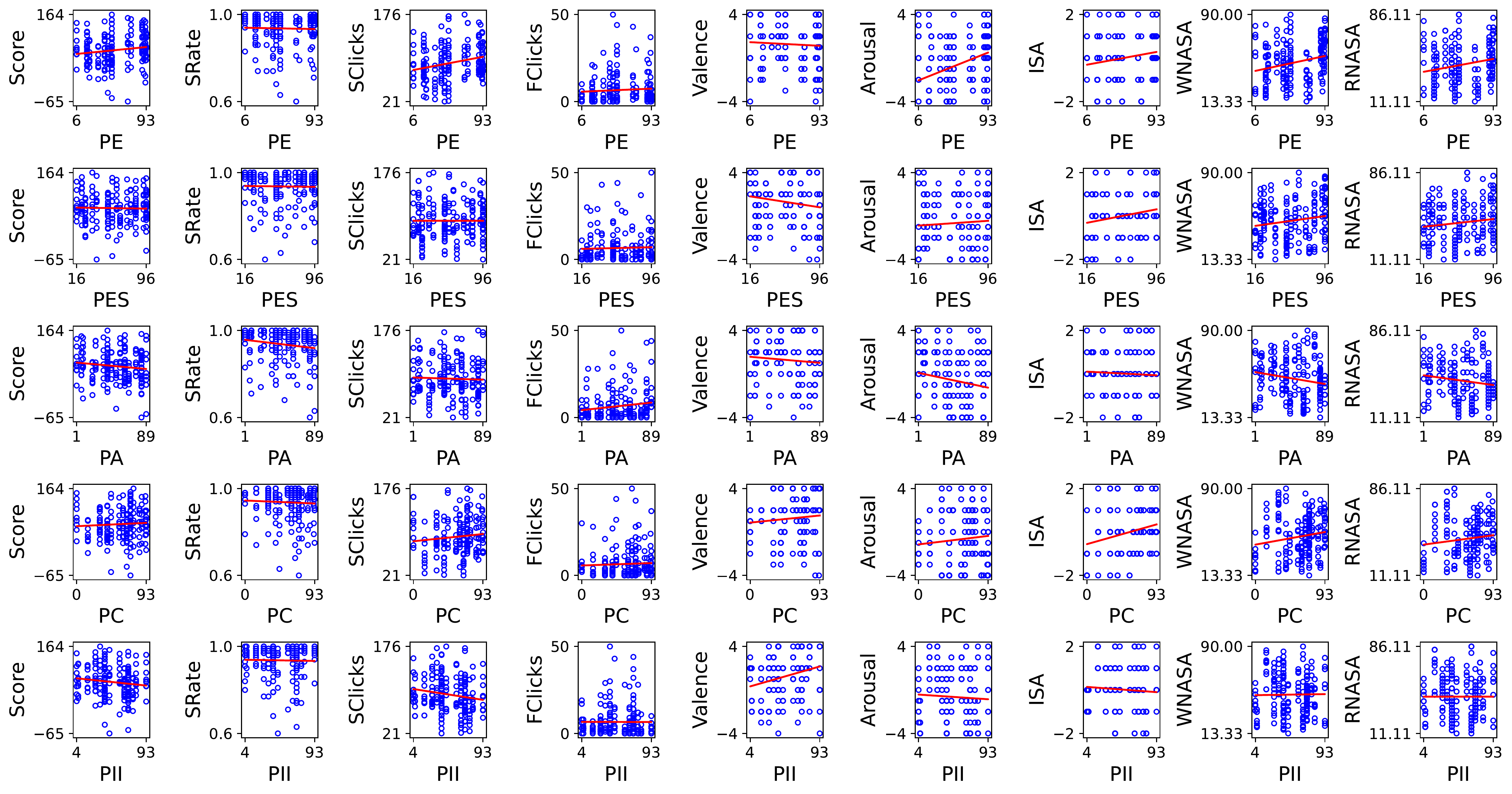}
    \caption{Scatterplots (in blue circles) and least-square linear fits (in red lines) of each pair of data between mission scores (Score), success rate (SRate), the number of successful clicks (SCliks), and failure ones (FClicks), valence, arousal, ISA, weighted NASA-TLX (WNASA), raw NASA-TLX (RNASA), PE, PES, PA, PC, and PII.}
    \label{fig:overall_personality_data}
    \vspace{-15pt}
\end{figure*}

We also investigate the data between personality traits, experimental data, and questionnaire results, as illustrated in  Fig.~\ref{fig:overall_personality_data}. Similar to Fig.~\ref{fig:overall_experimental_data}, the data with personality traits demonstrates multiple outliers. Moreover, no other significant statistical relationships are identified.

Table~\ref{tab:pearson_score1} and Table~\ref{tab:pearson_score} indicate that the experimental factors display more significant associations with performance, whereas the personality traits measure exhibit limited correlations. Additionally, Fig.~\ref{fig:overall_personality_data} shows that personality traits cannot solely account for the overall relevance of the experiment. Based on the results, we can deduce that the measured variables possibly have multivariate relationships; therefore, further investigation is warranted. 


\subsubsection{Exploratory Factor Analysis}

Although Pearson's correlation coefficient showed a weak correlation between personality traits and variables in the experiment, the method is limited because it can only consider two variables at a time. Also, the influence of personality traits on task performance had not been clearly identified. Furthermore, the dependency between personality traits and experimental variations is unclear from the perspectives of robot systems and cognitive processes; it can be argued oppositely. Therefore, we employed exploratory factor analysis (EFA) to investigate dependent and independent elements that could contribute as decisive factors in task performance. 

The EFA allows the measurement of relations in multivariate information with measured variables (observed variables) and categorizes them into latent variables (indirectly observed variables) \cite{ullman2012structural}. The purpose of the EFA is to examine the implications of quantified personality traits, experimental conditions, and perceived subjective ratings towards operators' performance measured through surveillance tasks. 

Among 187 data segments, we did not have missing data. Dimension reduction by varimax rotation was conducted to categorize measured values into latent variables except for experimental variables. The sampling adequacy value of KMO and Bartlett's test is .607 and the significance of Barlett's test of sphericity is less than .001, indicating that factor analysis is appropriate.

The selected variables are \textbf{Workload} (Weighted NASA\-TLX and ISA), \textbf{Performance} (Mission Scores and Success Rate), and \textbf{Personality} (PE, PES, PC). Other personality traits such as PA and PII did not show consistent results in grouping other human factors during EFA, and did not pass the KMO and Bartlett's test. The names of the latent variables were arbitrarily assigned to make them intuitive to group measured variables. Interestingly, the indicators of affective dimension, arousal, and valence, show low characteristics to be differentiated as a latent variable from varimax rotation; the arousal level was categorized as a potential variable with workload and the valence level was into performance level. 

\subsubsection{Structural Equation Modeling (SEM)}
\begin{figure}[t]
    \centering
    \includegraphics[width=0.88\linewidth]{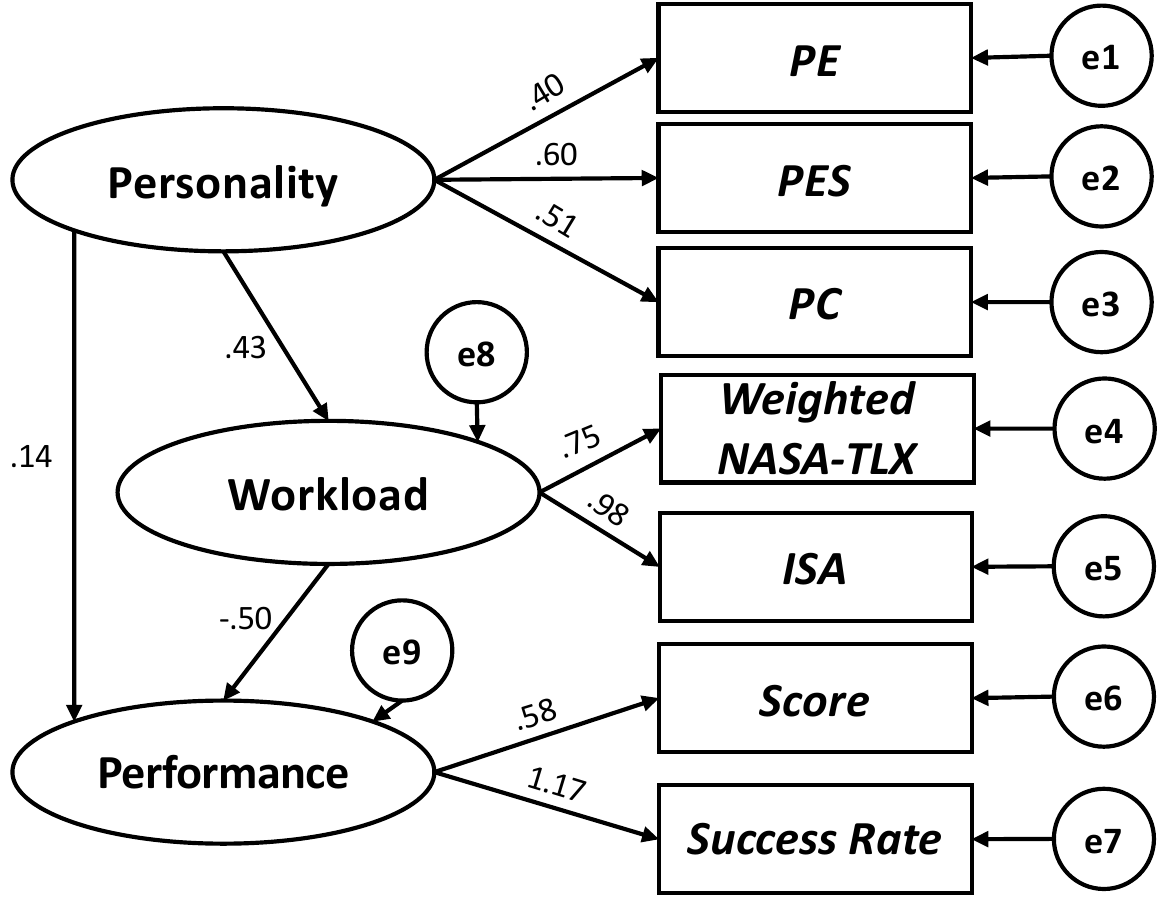}
    \caption{A path diagram in Structural Equation Modeling (SEM). Oval shapes represent latent variables, rectangular shapes represent measured variables. The circles with the caption starting with `e' indicate measurement errors in each observed variable. Solid straight lines refer to direct effects, and each number on the arrows indicates an individual standardized regression coefficient of each dependency.}
    \label{fig:sem}
    \vspace{-10pt}
\end{figure}

As shown in Fig.~\ref{fig:sem}, the SEM is created by the relationship between the aforementioned latent variables and measured variables into direct connections. The latent variables have been considered independent variables, although they showed associations in the previous section, to show the pertinence in multivariate analysis. Due to the direct connections of one latent variable to another one considered endogenous, the Personality variable and Workload have their errors as the same as measured variables. The numbers on each arrow indicate standardized regression coefficients.

The types of structural equation models require standard criteria to test the model fit with the root mean square error of approximation (RMSEA), comparative fit indices (TLI/CFI), and the significance of Chi-square (CMIN) \cite{fabrigar1999evaluating}. 
Although the discrepancy between studies was claimed, the acceptable fit of the SEM with less than 500 data should indicate the following values; RMSEA \textless 0.08, CFI/TLI \textgreater 0.90, and the significance of CMIN \textless 0.05 \cite{kline2015principles}. Additionally, the division of CMIN with the degrees of freedom (DF) shows a better fit if CMIN/DF \textless 3.0. The model in Fig.~\ref{fig:sem} indicates a moderate model fit by the values of RMSEA=$.064$, CFI=$.978$, TLI=$.957$, p-value=$.055$, and CMIN/DF=$1.758$.

\begin{table}[ht]
\caption{Estimates of Structural Model. }
\label{tab:regression_weight}
    \centering
    \resizebox{1\columnwidth}{!}{%
    \begin{tabular}{llll}
    \hline
    \multicolumn{1}{c}{\textbf{\begin{tabular}[c]{@{}c@{}}Endogenous \\ variables\end{tabular}}} &
    \multicolumn{1}{c}{\textbf{Estimate}} &
    \multicolumn{1}{c}{\textbf{\begin{tabular}[c]{@{}c@{}}Standardized \\ estimates\end{tabular}}} &
    \multicolumn{1}{c}{\textbf{p-value}} \\ \hline
Questions $\leftarrow$ Personality   & 0.36    & .012  & .003            \\
    Performance $\leftarrow$ Questions   & -12.372 & 3.715 & \textless{}.001 \\
    Performance $\leftarrow$ Personality & .293    & .206  & 154             \\ \hline
    \end{tabular}}
    \vspace{0pt}
\end{table}

\begin{table}[ht]
\caption{Total effect from measured variables to personality, performance, and workload.}
\label{tab:total_effect}
\begin{center}
\begin{tabular}{lccc}
\hline
                           & \textbf{Personality} & \textbf{Performance} & \textbf{Workload} \\ \hline
\textbf{PE}                & .514                & .000                & .000              \\ 
\textbf{PES}               & .595                & .000                 & .000              \\ 
\textbf{PC}                & .400                & .000                 & .000              \\ 
    \hline
\end{tabular}
\end{center}
 \vspace{-10pt}
\end{table}

Fig.~\ref{fig:sem} illustrates a latent variable that is dependent on and correlated with other variables. The SEM is used to confirm the existence of direct and indirect effects ($\mathcal{H}_4$) if personality has effectiveness indirectly between variables. As Table~\ref{tab:regression_weight} shows, regression weights, indicating individual relationships found, had their significance when the pairs of Personality $\times$ Workload and Questions $\times$ Performance. However, the pair of Personality $\times$ Performance might have a relationship, but it was rejected with a p-value of $.154$, which did not pass the significance test. 
As Table~\ref{tab:total_effect} shows, the total effect from measured variables is mainly related to their latent variables. The total effect is totally based on the direct effect, which means no indirect effect has not been found ($\mathcal{H}_4$; rejected). Each variable shows a consistent effect size on its latent variables. All personality indicators showed the total effect as 0, which means personality traits did not show their effect on performance measures in the experiment.

\section{Discussion} %
\label{sec:discussion}

\subsection{Personality Traits and Task Performance}
In Section~\ref{sec:analysis}, we conducted a correlation analysis, explanatory factor analysis, and structural equation modeling. These analyses indicate that not all personality factors necessarily affect performance in remote control situations when operators collaborate with multiple robots. While experimental variables have a substantial impact on the correlation between information perception and performance, personality traits exclusively have no effect ($\mathcal{H}_4$; rejected). However, this result is confined to the personality traits of extroversion, emotional stability, and conscientiousness. The total effects of agreeableness and imagination/intellect on task performance are unknown.

We found that extroversion had a positive impact on scores ($\mathcal{H}_1$a; accepted). Arousal and cognitive load were also found to be congruent with the assertion in \cite{humphreys1984personality}; the introversion-extroversion is the potential to elicit performance levels. Additionally, the total effects in the factor analysis were entirely comprised of direct effects, indicating their insignificant effect. The degree of introversion-extroversion may not show its effectiveness in achieving better performance, which is lower than expected. Therefore, we can conclude that the degree of extroversion in surveillance tasks does potentially require motivation or direction and intensity depicted in \cite{humphreys1984personality}. 

We also found that agreeableness had an overall negative effect, especially on click success rate, although the negative inclination on scores cannot be reproducible in high probability  ($\mathcal{H}_2$b; partially accepted). According to the definition of PA, participants with higher agreeableness may have an optimistic inclination, leading to prompt decisions to click their response to incorrect objects, as supported by the evidence in \cite{haas2015agreeableness}; a shorter time latency to discern visual stimuli is faster when participants indicate higher agreeableness. However, the interpretation of this result is challenging to derive a conclusion because the necessity of affective load was not substantially emphasized. 

Lastly, the dimension of conscientiousness was found to have weak relevance with task performance and completion simultaneously ($\mathcal{H}_1$b, $\mathcal{H}_2$a; rejected). Previous research suggested that people with higher conscientiousness showed better performance in engaging environments and showed a lower rate of recognizing errors \cite{hill2016contextualizing, bakker2012work}. However, the outcome is not compatible with the studies, which may be supposedly caused by situational moderators \cite{humphreys1984personality}. Considering the experiment asked the participants to click defined objects streamed with different numbers and speeds, we should notice a major difference. Click events were made through a mouse that should navigate the CCTV-formatted GUI which led to asking participants physically move their hands. People with higher conscientiousness are believed to be self-regulated, which shows the same tendency in initiating activities \cite{wilson2015personality}. This means individuals of high conscientiousness potentially suppress their gestures and behavior even though they could recognize abnormal objects.

\subsection{Personality Traits and Cognitive Workload} 
The dimension of extroversion, emotional stability, and conscientiousness was found to have significant associations with cognitive workload indicators. Arousal was found to have limited implications on information processing, as arousal was evaluated as a subordinate factor to the workload latent variable during the Varimax rotation analysis. The evaluation of workload using the NASA-TLX questionnaire or self-reported workload with ISA showed positive associations with the higher three personality traits, merging with the results of affective measures. This suggests that participants with these personality traits may solely occupy their perception ability in discerning the true or false visually. 

The level of arousal was found to be positively associated with extroversion and negatively associated with agreeableness. According to \cite{humphreys1984personality}, extroverts are expected to accept more arousal than introverts in terms of information transfer, potentially eliciting prompt behavioral responses to identify desired input in experimental GUI. Conversely, when recipients of information showed lower agreeableness, their arousal decreased. 
The correlation $\gamma$ of PA also showed a decreasing tendency despite its insignificance, suggesting that arousal level may also be one of the factors to decide the score. However, this conclusion is based solely on the $\gamma$ value, which is calculated from scattered data.


\subsection{Personality Traits and Valence}
Valence is the least significant variable among the measured variables. There appears to be no statistically significant influence on affect when collaboration between humans and robots performs task-specific interactions. During the Varimax rotation, valence was mostly grouped with performance-related measurements, which suggests that higher overall performance may result in higher valence. The tendency between measured data shows the lacked necessity of affect could be connected to the successful clicks leading to attaining a positive level of pleasure.


Emotional stability showed a negative correlation with valence, but a positive correlation with cognitive load, although the significance levels were different. In particular, valence tended to be slightly higher when emotional stability was lower, indicating a weak relationship. This relationship could be due to the task not requiring emotional load, as was assumed with the agreeableness indicator. 

The Imagination/Intellect index only displayed statistical significance in relation to valence, showing a small positive correlation. This suggests that the experiment was more positively received when participants had a higher index score. Considering that the index is occasionally interpreted as `openness to experience', it is conceivable that the experiment was accepted as a new experience, which may have influenced the results. 

For our analysis, we utilized the data obtained by measuring the cognitive load, affect, and performance of 21 subjects exposed to nine experiments under different conditions. While the analysis was based on the sample size of 21 individuals, the results suggest that there may be a generalizable relationship between personality traits and task performance. Further research could investigate this relationship across a broader range of personality distributions.

\section{Conclusion and Future work}
\label{sec:conclusion}
In this paper, we investigated the implications of personality trait indicators on cognitive workload, affect, and task performance in the context of remote robot control using the open-access multi-modal dataset MOCAS. Hypotheses were tested to explore the links between performance and behavioral/affective ratings based on users' personality traits. Based on our analysis, we concluded that human personality traits should conditionally be regarded when operating robots, especially if the goal is to improve the task completion rate. Personal characteristics that are not confined to personality also play a critical role in human-robot teams. Although the analyzed human personality characteristics may seem minimal, the effect of personality traits is not negligible, as demonstrated in diverse fields. Therefore, it is essential to give the necessary consideration to individual differences when designing human-robot teams.

Future research could investigate whether a combination of personality traits can lead to even more successful task completion when operators collaborate. Additionally, it would be useful to investigate which intrinsic attributes of users should be considered more in an adaptive system.


\bibliography{root}
\bibliographystyle{IEEEtran}

\end{document}